\documentclass[letterpaper, 10 pt, conference]{ieeeconf}  

\IEEEoverridecommandlockouts                              

\usepackage{graphicx} 
\usepackage{fancyhdr}
\fancypagestyle{withfooter}{
  
  \fancyfoot[C]{\footnotesize Accepted to the IEEE ICRA Workshop on Field Robotics 2024}
}

\title{\LARGE \bf
Multi-purpose robot for rehabilitation of small diameter water pipes
}

\author{Julien~Feiguel$^{1}$, Mouhamed~NDiaye$^{1}$, Pascal~Chambaud$^{1}$, Adrien~Chambellan$^{1}$, Pierre~Blanc$^{1}$,  Steve~Bourgeois$^{1}$,\\
 Lucas~Labarussiat$^{1}$,  Clémence~Dubois$^{1}$, Audrey~Vigneron$^{1}$, Thomas~Desrez$^{1}$, Alain~Riwan$^{1}$, and~Caroline~Vienne$^{1}$
\thanks{$^{1}$Université Paris Saclay, CEA, LIST,
        Palaiseau, France
        {\tt\small firstname.lastname@cea.fr}}%
}

\begin{document}

\maketitle
\thispagestyle{withfooter}
\pagestyle{withfooter}

\begin{abstract}

Rehabilitating cast iron pipes through lining offers several advantages, including increased durability, reduced water leaks, and minimal disruption. This approach presents a cost-effective and environmentally-friendly solution by sealing cracks and joints, extending the pipeline's lifespan, and reducing water wastage, all while avoiding the need for trench excavation. However, due to the relining process, branch connections are sealed and need to be reestablished. To address the issue of rehabilitating small-diameter water pipes, we have designed a modular robot capable of traversing and working within 200-meter-long, 100mm-diameter cast iron pipes. This robot is equipped with perception functions to detect, locate, and characterize the branch connections in cast iron pipes and relocate them after lining, as well as machining functions. A first prototype of this system has been developed and validated on an 8-meter-long section, in a laboratory environment.

\end{abstract}

\section{INTRODUCTION}

The integrity and efficiency of drinking water distribution systems are critical to environmental sustainability and economic stability. A significant challenge facing these infrastructures globally is the deterioration leading to leaks, with some networks reporting as much as 20\% water loss. This not only results in substantial resource wastage but also increases the financial burden on utilities and consumers alike. Traditional pipeline rehabilitation methods, involving extensive trench-digging, disrupt local communities and landscapes, and incur high costs due to the excavation and displacement of large volumes of soil. In response, trenchless rehabilitation technologies have gained prominence due to their less invasive nature and reduced environmental impact.

One effective trenchless technique involves the insertion of a high-density polyethylene (HDPE) liner inside existing cast iron pipes, which seals leaks and restores structural integrity. This method, typically applied to straight pipeline sections up to 200 meters in length, requires only access pits at either end of the segment for liner insertion. However, this retubing process seals off branch connections to customer properties, necessitating subsequent interventions to re-establish these essential service connections.

To address this issue, we propose a novel robotic solution in collaboration with SADE, aiming to realize the restoration of branch connections post-retubing from inside the HDPE tube. 
The adopted strategy involves a first pass through the bare cast iron to locate and map the branch connections through visual perception as well as characterize their state. During this pass, the branch connections will be re-bored to ensure a shape and surface condition suitable for the future insertion of a sealing component. A second pass is made, after relining, to pierce the PE precisely at the branch connection location, and finally, a third pass will be dedicated to the restoration of the branch connections. This last operation is not handled by the current robotic system but will be addressed in future developments.

The large majority of the literature of in-pipe robot development adresses inspection task~\cite{c1} or cleaning tasks in large diameters pipe~\cite{c2}. To our knowledge, there is no other robotic system capable of carrying out detection, characterization and machining operations in such narrow pipes. The contributions that we highlight therefore concern the general design of the solution and the mechanical and electrical realization of the system as well as the integration of advanced perception functions by vision and eddy currents.

\section{Prototype design}

\subsection{Rehabilitation concept and design requirements}

The water pipes we aim to rehabilitate have a nominal diameter of 100 mm, but this diameter is reduced to 80 mm after the installation of the HDPE pipe. In order to incorporate all the necessary movement, perception, and action functionalities required for the rehabilitation of these water pipes from within, while adhering to the constraint of an 80 mm diameter, the cylindrical-shaped robotic system reaches a length of approximately 2 meters. However, due to the slight angle between different segments of the pipeline, it is not feasible to create a monolithic robot exceeding 2 meters in length with a diameter close to the 80 mm limit. To ensure passage through the junctions, a design with multiple wagon-like units articulated together is necessary. 

The robot is thus composed of various modules, all having an outside diameter of 75~mm to fit both within the cast-iron and HDPE pipes:
\begin{itemize}
\item A traction module equipped with two motors for propulsion and clamping within the tube, along with two variators.
\item Electronic assembly made up of several sub-modules
\begin{itemize}
\item an high-speed optical network module for long-distance communication inside the tube 
\item an embedded IPC to manage the robot's actuators and sensors 
\item an embedded computer to manage the video stream from the cameras
\item Eddy current acquisition electronics
\item voltage conversion sub-modules for the various systems
\end{itemize}
\item A controller and conversion card module for controlling the motors of the modules.
\item A final interchangeable operational module, which is respectively used to bore the cast iron (first pass) and to drill the HDPE (second pass).
\end{itemize}

The mechatronic system is powered by an external power source of 230Vac - 16A. The control station is located outside the pipeline. An umbilical tether provides power to the robot and fiber optic communication facilitates digital data exchange. We ensure the effective coordination of the perception, locomotion, and action functionalities through ROS2 middleware~\cite{c3}. This middleware facilitates the communication by managing data streams while adhering to a common clock for temporal synchronization of data. Additionally, it has the ability to abstract the underlying hardware architecture, thus enabling the system to run on one or multiple computers operating on heterogeneous operating systems.

\subsection{Locomotion}
\begin{figure}[thpb]
	\centering
	\includegraphics[width=0.5\textwidth]{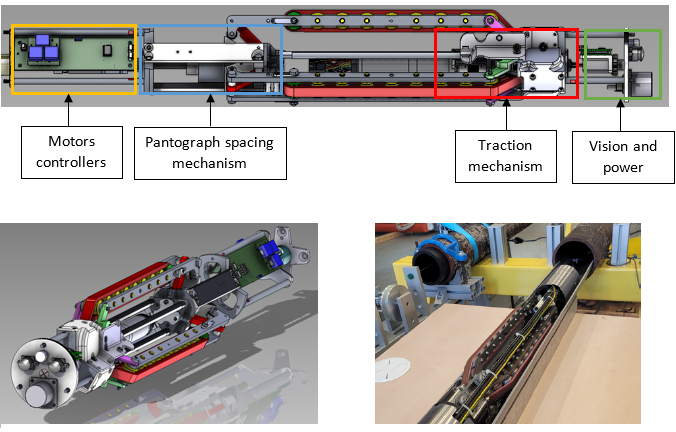}
	\caption{Locomotion function of the robot (CAD model and picture).}
	\label{fig:loco}
\end{figure}

The traction module is necessary to move the entire system through the 200 m section of the pipe. It can be separated into four distinct parts (see Fig.~\ref{fig:loco}):
\begin{itemize}
\item The vision system, which will be detailed subsequently.
\item A pantograph spacing mechanism, which is powered by a motor combined with a 3x120° wedge screw mechanism. This mechanism pushes the parallelograms upwards, enabling the traction module to remain concentric to the pipes within a wide range of diameters, starting from 75 mm to 120 mm. It exerts a force of up to 300 N.
\item The traction mechanism itself utilizes tracks to navigate the robot through the 200 m section. It is driven by a 100 W motoreducer connected to a 45° helix angle worm screw, which in turn drives three worm wheels at 120° each to power individual tracks. This configuration allows the robot to generate a traction force of 370 N at a speed of 10 cm/s or 510 N at 5 cm/s by adjusting the screw gear ratio.
\item The motor controllers that are positioned at the back end of the traction module.
\end{itemize}

\subsection{Machining}
\begin{figure}[thpb]
	\centering
	\includegraphics[width=0.23\textwidth]{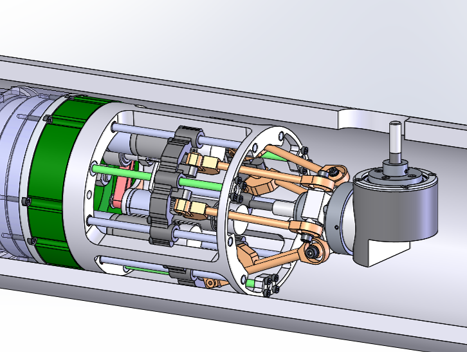}
	\includegraphics[width=0.23\textwidth]{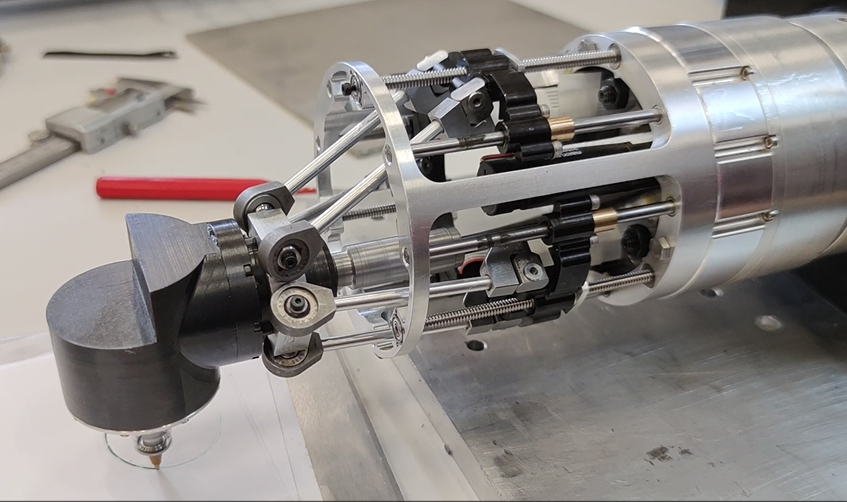}
	\caption{Machining function of the robot (CAD model and picture). On the right, the delta mechanism integrates a pen to visualize and validate the trajectory.}
	\label{fig:delta}
\end{figure}

Since the branch connections of the pipe can be situated on the sides or on the top, the operational module must be adaptable to each scenario. Consequently, it is linked to a 400° rotating and locking mechanism, ensuring concentricity within the pipe. This enables scanning of the various holes to determine their relative center and axis before initiating machining (refer to Fig.~\ref{fig:delta}). Three translations are necessary to perform the machining operation. We selected a delta-type robotic kinematics, which provides high stiffness and workspace within the harsh geometrical constraints.

The machining module is composed of three distinct parts:
\begin{itemize}
\item The delta mechanism powered by three 8W motors connected to ballscrews and linear guides to maintain a sufficient rigidity while machining provides up to 350N/ screw in continuous mode. The mechanism is totally backlash free to maintain the machining precision as high as possible. 

\item The machining head employs an 80 W motor connected to a spur gear to maximize torque while maintaining a suitable machining speed. Tools can be interchanged using a 6 mm lathe collet. For PE, a larger diameter tool is utilized to expand the range of hole diameters the machine can handle. Depending on the material, the machining head can sustain speeds of up to 8000 rpm and torque of 0.5 Nm, hence cutting depths are kept shallow to minimize radial loads.

\item Sensors and electronics to detect precisely the center of the holes using vision, laser and Eddy current probe.
\end{itemize}

\subsection{Sensors}
While navigating in the cast pipe, visual perception is used to determine the location of the robot through a simultaneous localization and mapping (SLAM) approach~\cite{c4} and to build a map of the branch connections. To carry out these two tasks, a front vision system is made up of three elements (see Fig.~\ref{fig:vision}): 
\begin{itemize}
\item A laser allowing detection of branch connections 
\item A pair of cameras with a wide field of view and depth of field adapted to the use case, along with cable connectors, allowing both the electronic board to be separated from the cameras and providing freedom to choose the parallax.
\item Lighting for illuminating the scene so that the cameras can observe the texture of the pipes, allowing the SLAM algorithm to utilize this information to estimate the 3D movement of the robot
\end{itemize}

\begin{figure}[thpb]
	\centering
	\includegraphics[width=0.33\textwidth]{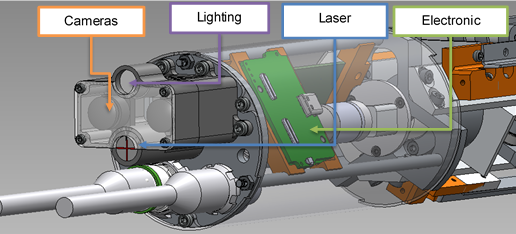}
	\includegraphics[width=0.13\textwidth]{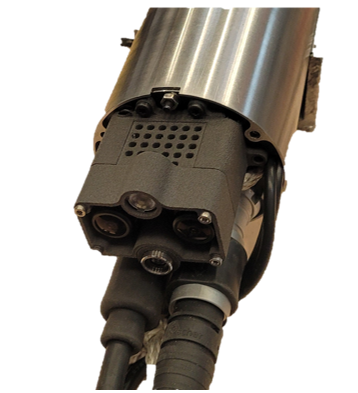}
	\caption{Integration of the front vision system: CAD model (left) and picture (right).}
	\label{fig:vision}
\end{figure}

In addition to this front vision system, an additional back vision system is integrated in order to characterize the state of the branch connections. Its objective is twofold: first, to identify non-compliance cases that do not allow for a branch repair from the inside (significant offset between the branch axis and the valve), and second, to accurately locate the center of this branch to enable precise positioning of the drill for machining.
The proposed solution leverages the robot's ability to rotate around its axis to perform a 3D reconstruction through profilometry, i.e., by projecting a laser stripe whose deformation observed by a camera allows deduction of the 3D shape. 

Once the internal lining step has been completed, the robot can no longer use the vision modality to detect the holes. A non-destructive testing modality allowing detection through HDPE has therefore been proposed. This measurement system, based on Eddy Currents, includes two axial coils dedicated to locate the connections in the axis of the pipeline through differential measurement and an additional coil dedicated to identifying the center of the connection. More details are given in the following section.

\section{Methodology and results}
In this section, we specifically describe the methodologies of perception through front vision and through eddy currents implemented to locate the branches during the advancement of the robot.

\subsection{Detection and localization of branch connections through vision} 
\begin{figure}[thpb]
  \begin{tabular}{cccc}
            \includegraphics[width=0.1\textwidth]{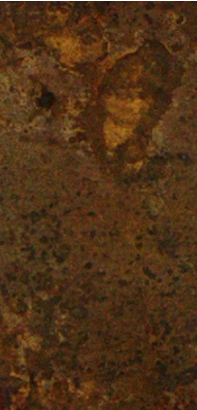} & 
            \includegraphics[width=0.1\textwidth]{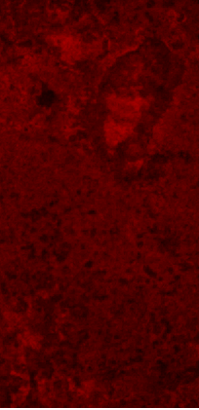} & 
            \includegraphics[width=0.1\textwidth]{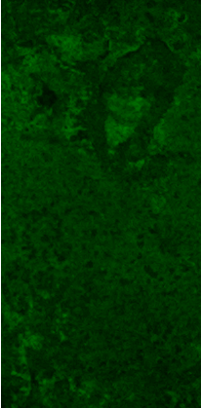} & 
            \includegraphics[width=0.1\textwidth]{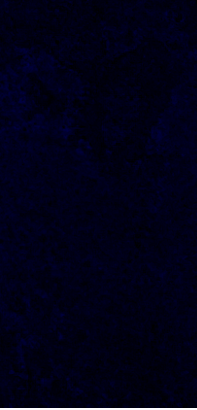} \\
             color & red & green & blue \\
   \end{tabular}
	\caption{Observation of a sample of cast iron pipe across all channels and each channel separately.}
	\label{fig:channels}
\end{figure}

As mentioned earlier, the front vision system is used both for robot localization using SLAM and for branch detection and location based on laser input. To prevent interference between these tasks, we propose projecting light in different spectral domains so that each light projection does not affect the same color channels of the camera. To determine if this option of spectral separation is functional, we observed the spectral response of a cast iron pipe for the different spectral bands of a color camera. To do this, we illuminated the pipe with a perfectly white light source and observed the light intensities for each of the three channels (red, green, and blue) of the camera. As illustrated in Figure~\ref{fig:channels}, it appears that the spectral response of the cast iron pipe mainly covers the red and green domains, whereas the response in the blue domain is significantly weaker. The lack of contrast in the blue channel compared to the red and green channels indicates that it will have a negligible contribution to the detection and characterization of visual landmarks used by the SLAM. The similarity of the red and green channels indicates that they provide similar information, and the use of either should be equivalent for the SLAM algorithm.

Based on this analysis, the chosen approach involves using a red laser for branch detection and lighting projecting in the green spectral domain for SLAM.

An initial experiment is conducted to assess the system's capability to estimate the distance between consecutive branches. The results presented (see Tab.~\ref{Tab:Tcr}) compare the actual distances (measured by a human operator) with those derived from the 3D point cloud reconstructed by the SLAM algorithm. 

\begin{table}[ht]
\caption{Measured minimal and maximal errors in position estimation through vision on six different pipe sections.}
\centering
   \begin{tabular}{| l || p{1cm} | p{1cm} | p{1cm} | p{1cm} | }
     \hline
      & Min error (mm) & Min error (\%) & Max error (mm) & Max error (\%) \\ \hline
     S1 (0.23m) & 1 & 0.5\% & 4 & 0.19\% \\ \hline
     S2 (0.308m)  & 2 & 0.6\% & 5 & 0.16\% \\ \hline
     S3 (0.265m)  & 6 & 2.5\% & 6 & 2.5\% \\ \hline
     S4 (0.63m) & 5 & 0.8\% & 14 & 2.2\% \\ \hline
     S5 (1.635m) & 0 & 0.0\% & 33 & 2\% \\ \hline
     S6 (1.50m) & 8 & 0.5\% & 8 & 0.5\%  \\
     \hline
   \end{tabular}
   \label{Tab:Tcr}
 \end{table}

The experiments conducted revealed that the branch connections aligned with the laser were all successfully detected (see Fig.~\ref{fig:vision}). The maximum observed error reached a peak of 2.5\% of the distance between two branches and reached an absolute maximum of 3.3 cm on the longest section. Due to the methodology employed, the detected position primarily corresponds to the intersection of the rear wall of the branch connection with the laser, rather than its center. Nonetheless, this location is generally sufficient as it allows for indicating the position of the branch connection within a few centimeters, although further refinement of the position with the rear system is required in all cases.

\begin{figure}[thpb]
	\centering
	\includegraphics[width=0.48\textwidth]{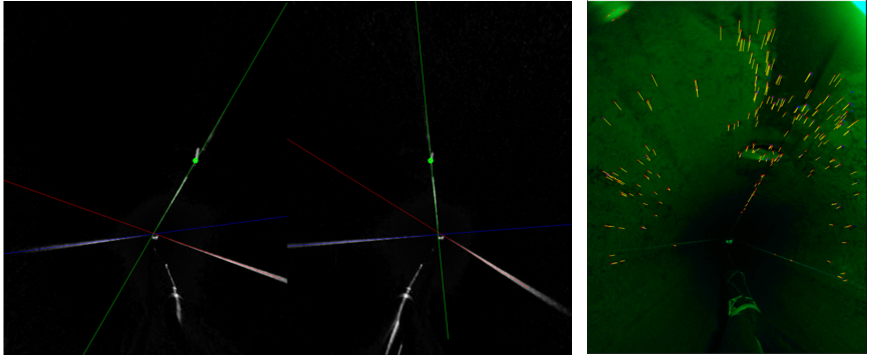}
	\caption{Example of branch detection. Left: the two images of the lasers (red channel of each video stream) with the detection of the laser lines (colored lines) and the detection of the branch connection (green circle). Right: image of the green channel of the left video stream with features detection using SLAM approach.}
	\label{fig:vision}
\end{figure}

\subsection{Characterization of branch connections through profilometry} 
\begin{figure}[thpb]
	\centering
	\includegraphics[width=0.49\textwidth]{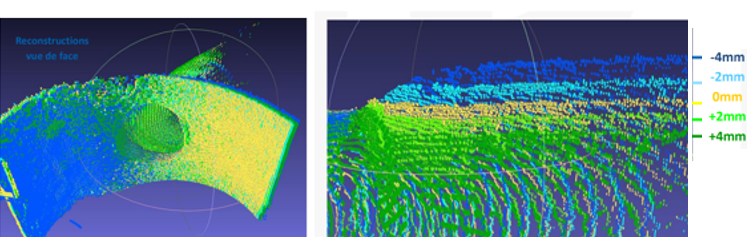}
	\caption{Example of a combination of shifted 3D reconstructions (on the left), along with a zoom on the part corresponding to the branch (on the right). An error of 0.35 mm is observed between the most extreme reconstructions.}
	\label{fig:profilo}
\end{figure}

After initially positioning the robot roughly using its front vision capability, the characterization of the branch connection is performed through a profilometry approach embedded in the final module. The objective is to reconstruct the point cloud of the branch to confirm its suitability for robotic intervention and to precisely locate its center. To assess the performance of this perception task, several reconstructions of the same branch are conducted by varying the position of the valve relative to the branch connection and the position of the machining head relative to the branch. As there is no ground truth regarding the orientation of the branches, this evaluation relies on assessing the diameter of the branch connection and the valve. In the current configuration, the time required to complete the reconstruction is approximately 10 minutes for a 360° scan with an angular step of 1°. However, it should be noted that a 360° scan is not necessary since the position of the branch is approximately known during the characterization step. Therefore, this time can be significantly reduced.
The results obtained are illustrated in Figure~\ref{fig:profilo}. The measurements taken indicate an average reconstruction error of approximately 0.3 mm.

\subsection{Eddy current perception}
To relocate the branch connections across the HDPE tube, the robot is equipped with coils and sensors allowing for Eddy Current measurement. The principle of this method, used notably in Non-Destructive Testing, involves supplying a coil with a sinusoidal electrical current to create a magnetic field. In the presence of a conductive material, so-called Eddy currents are induced, and a reactive magnetic field is created in turn. In our case, the presence of holes in the cast iron at the branch connections induces a variation in the electrical impedance of the coil, which is automatically detected and used to position the robot for machining the HDPE.

The localization process consists of two steps using two distinct probes. The first probe is an axial probe, constituted of two coils used in a differential mode, which provides an axial localization of the branch connection center. Then, a point coil mounted on the last operational module of the robot, is positioned at the axial position provided by the first probe. A scan is performed along the axis and circumference of the pipeline to determine the radial position of the branch connection as well as the precise axial position since the initial axial detection may suffer from robot positioning errors.

\begin{figure}[thpb]
	\centering
	\includegraphics[width=0.23\textwidth]{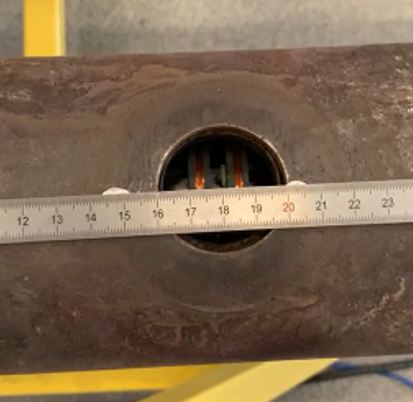}
	\includegraphics[width=0.18\textwidth]{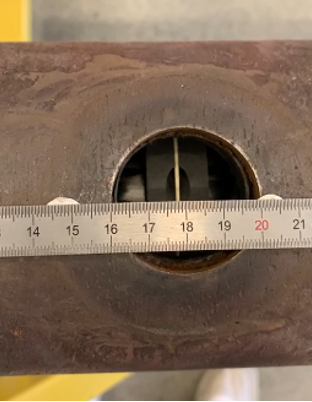}
	\caption{View of the tests validating the performance of the branch detection through Eddy current.}
	\label{fig:eddy}
\end{figure}

The acquisition of data from the measurement system and the position of the robot is carried out by a system synchronized to the microsecond. The function is executed within the expected tolerances on two consecutive branch connections without impact of the presence or absence of the connection itself. In Fig.~\ref{fig:eddy}, we see in particular on the same hole the passage of the coils triggering the stop (left) and the stabilization of a marker representing the position of the radial coil in the center of the tap (right).

\begin{figure}[thpb]
	\centering
	\includegraphics[width=0.49\textwidth]{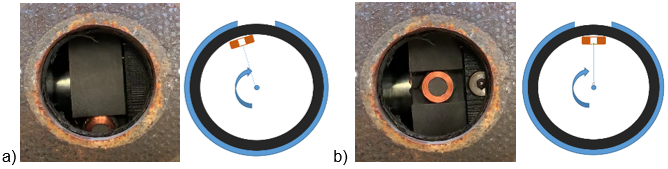}
	\caption{Photo and diagram of the coil offset (a) and aligned (b) with the hole in the pipe.}
	\label{fig:coil}
\end{figure}

The Figure~\ref{fig:coil} illustrates two positions a) and b) of the radial coil during this phase, associating each of the captured photos with a diagram indicating the corresponding position of the coil. In the diagram, the outer blue circle represents the cast iron pipe, the interrupted area corresponding to the branch connection, and the inner black circle represents the HDPE tube. The white and orange rectangle represents the coil, offset from the hole of the branch connection in the left configuration and aligned with it in the right configuration. Finally, the blue arrow represents the rotational movement performed by the probe.

\begin{figure}[thpb]
	\centering
	\includegraphics[width=0.5\textwidth]{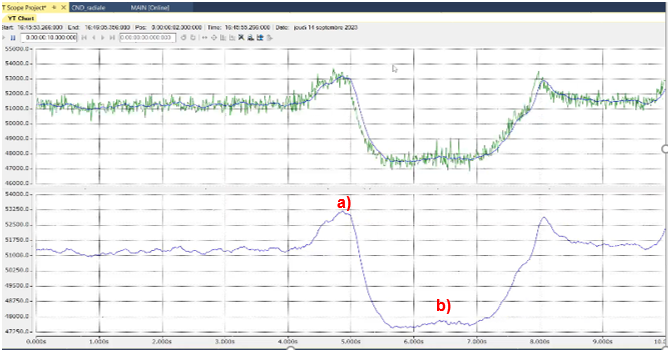}
	\caption{Display of the raw (green) and filtered (blue) measured EC signal as a function of the time.}
	\label{fig:curve}
\end{figure}

Positions a) and b) are also identified in Figure~\ref{fig:curve}, which represents the measured signal. This signal (blue curve) corresponds to the modulus calculated from the real and imaginary parts of the complex impedance of the coil (in green), filtered by a moving average over 15 values. These real and imaginary parts are obtained from the demodulation of the digitized signal. The demodulation frequency corresponds to the excitation frequency. The signal amplitude (y-axis) is displayed against time (x-axis), expressed in seconds. The hole position is finally determined from the information about the robot's speed of movement within the pipeline.

The signal processing from the radial coil ensures the radial centering of this coil on the hole. For radial measurement, a balancing phase then amplification of the signal before digitization is carried out in order to maximize the measurement response.

\subsection{Prototype validation}
The different functionalities of the robotic system are tested sequentially in two 8-meter pipes installed in the laboratory. For the system's first pass, we have validated the ability of the system to detect, locate, and characterize a branch of the cast-iron pipe as well as the ability of the operational module to bore a preformed hole. The total time for the machining operation is around 30 mn to enlarge a hole from a diameter of 20 mm to 24.4 mm. 

\begin{figure}[thpb]
	\centering
	\includegraphics[width=0.23\textwidth]{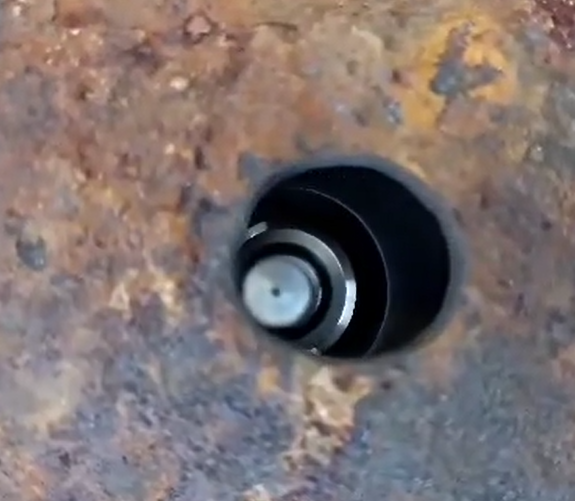}
	\includegraphics[width=0.23\textwidth]{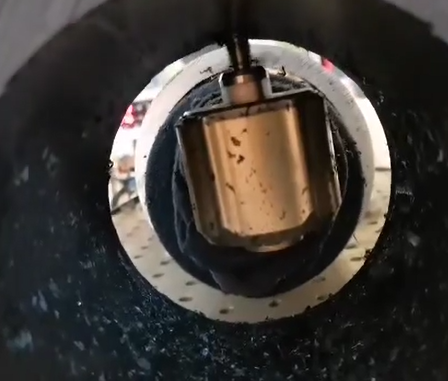}
	\caption{Machining tests are realized inside the pipe on cast iron (left) and PE (right).}
	\label{fig:machining}
\end{figure}

The robot is then inserted into the second pipe, which has been relined with a HDPE tube with an internal diameter of 80 mm. The tested sequence then involves detecting the branches using the axial probe for coarse pre-positioning of the robot, and then conducting a fine mapping to precisely position the drill at the center of the branch. Finally, a PE machining test is performed (see Fig.~\ref{fig:machining}).

The machining of the PE is performed using a Ø 20mm milling cutter with Z=2. However, first tests demonstrated that this tool is suboptimal due to the low number of teeth, which results in a high helix angle. Consequently, there is a tendency for the cutter to screw and jam into the material if the rotational speed is not high enough. To mitigate the occurrence of cutter jamming during drilling, we vary the spindle speed between 2800 and 3500 rpm, instead of the recommended 1000-1300 rpm for plastics. This higher speed helps counteract screwing effects caused by tool inertia but may lead to local material melting, thereby increasing viscosity. However, this phenomenon primarily occurs during drilling, the operation that demands the most effort. Drilling remains the most delicate operation. By adjusting the spindle rotation speed parameters, we can achieve drilling in approximately 9 minutes, followed by reaming up to a diameter of 23 mm, with a machining time of about 4 minutes.


\section{Conclusion and Future works}
Up until now, we have tested the robot's ability to detect, characterize and bore branches. This involves the collaboration of vision functionalities (branch detection, motion estimation, characterization), locomotion (motor control for positioning in characterization/machining position, control of machining head movement for characterization) and the machining itself. After PE retubing, we have assessed the robot's ability to locate a branch through the PE and to machine the PE. This involves the collaboration of Eddy current functionalities (axial and radial) with locomotion (motor control) and PE machining. In future development, we will have to assess quantitatively the performances in terms of precision of positioning, duration of the different robotic operations, and quality of the machining. In parallel, further validation is planned through field tests to validate the performance and robustness of the solution on a 100-meter section of pipeline before and after retubing.


\section*{ACKNOWLEDGMENT}

Part of this work is supported by SADE, our industrial partner in this project.



\section*{APPENDIX}

\begin{figure}[thpb]
	\centering
	\includegraphics[width=0.53\textwidth]{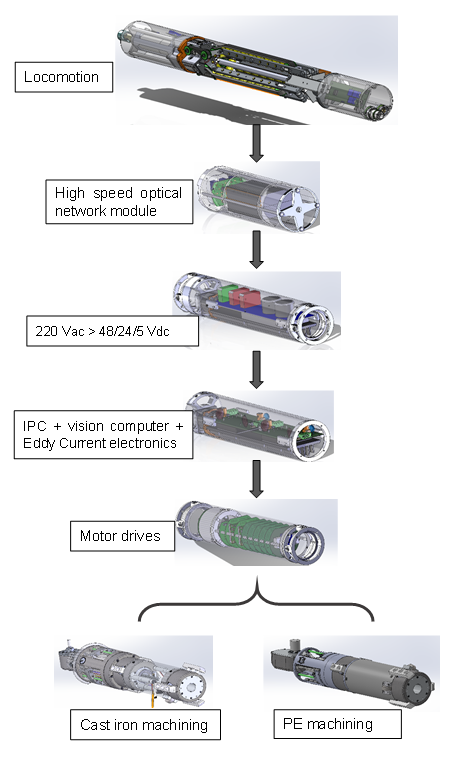}
	\caption{General overview of the proposed system for rehabilitation of small diameters water pipes.}
	\label{fig:all}
\end{figure}

In this section we display the general overview of the robotic system (see Fig.~\ref{fig:all}).

\end{document}